\documentclass{article}

\usepackage[preprint]{neurips_2026}

\usepackage[utf8]{inputenc}
\usepackage[T1]{fontenc}
\usepackage{hyperref}
\usepackage{url}
\usepackage{booktabs}
\usepackage{amsfonts}
\usepackage{amsmath}
\usepackage{amssymb}
\usepackage{graphicx}
\usepackage{microtype}
\usepackage{xcolor}
\usepackage{enumitem}
\usepackage{placeins}

\newcommand{\goodmark}{\raisebox{-0.18em}{\includegraphics[height=1.1em]{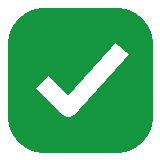}}}
\newcommand{\badmark}{\raisebox{-0.18em}{\includegraphics[height=1.1em]{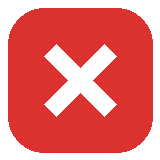}}}
\newcommand{\zhengmeid}{\raisebox{-0.18em}{\includegraphics[height=1.15em]{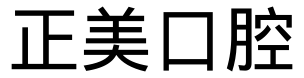}}}
\newcommand{\huatianhua}{\raisebox{-0.18em}{\includegraphics[height=1.15em]{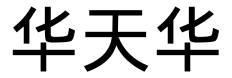}}}
\newcommand{\beijingduck}{\raisebox{-0.18em}{\includegraphics[height=1.15em]{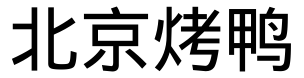}}}

\title{Allegory of the Cave: Measurement-Grounded Vision-Language Learning}
\author{%
  Kepeng Xu, Li Xu, Gang He, Wenxin Yu\\
  ghe@xidian.edu.cn\\
  Xidian University\\
  Southwest University of Science and Technology
}

\begin{document}

\maketitle

\begin{abstract}
Vision-language models are almost universally trained and evaluated on post-ISP RGB images, implicitly treating rendered appearance as a sufficient interface for multimodal grounding. However, RGB rendering is a lossy observation of the underlying sensor measurement: in low-light, high-dynamic-range, and exposure-imbalanced scenes, image signal processing will clip highlights, suppress structures, quantize evidence, and discard task-critical visual signals before reasoning begins. We study whether VLM grounding improves when the model observes a measurement-domain representation that preserves richer sensor evidence than rendered RGB. To this end, we formulate \emph{measurement-grounded vision-language learning} and instantiate it as \emph{PRISM-VL}, a framework that adapts vision-language models to RAW-derived measurement-domain inputs. PRISM-VL combines three design choices: a linear measurement-domain input that preserves sensor-proximal signal, camera-conditioned grounding through metadata-augmented questions and residual metadata conditioning in the visual encoder, and Exposure-Bracketed Supervision Aggregation (BracketSup), which uses exposure-conditioned RGB proxies for annotation while attaching supervision to the underlying measurement-domain capture. We construct a quality-controlled 150K instruction-tuning resource and a held-out benchmark targeting low-light, HDR, visibility-sensitive, and hallucination-sensitive cases.  PRISM-VL improves grounding accuracy over RGB baselines, reaching 0.6120 BLEU, 0.4571 ROUGE-L, and 82.66\% LLM-Judge accuracy, corresponding to gains of +0.1074 BLEU, +0.1071 ROUGE-L, and +4.46 percentage points over the RGB Qwen3-VL-8B baseline. These results indicate that part of VLM grounding error originates from information lost during RGB rendering, and that preserving measurement-domain signal can provide more complete evidence for accurate multimodal reasoning.
\end{abstract}

\section{Introduction}

Post-ISP RGB has become the default visual interface for vision-language models (VLMs). A camera, however, does not observe RGB directly: it records sensor measurements that are later transformed into a display-oriented image through demosaicing, white balancing, denoising, tone mapping, clipping, and quantization. This rendering pipeline is useful for human viewing and indispensable for standard vision workflows, but it is not neutral with respect to evidence preservation.

\begin{figure*}[t]
\centering
\setlength{\tabcolsep}{3pt}
\renewcommand{\arraystretch}{1.05}
\resizebox{\textwidth}{!}{%
\begin{tabular}{c c c c c}
\toprule
Meas.-XYZ image & RGB image & Lost-Signal & Lost-Signal Crop & Luminance loss distribution \\
\midrule
\includegraphics[width=0.16\textwidth]{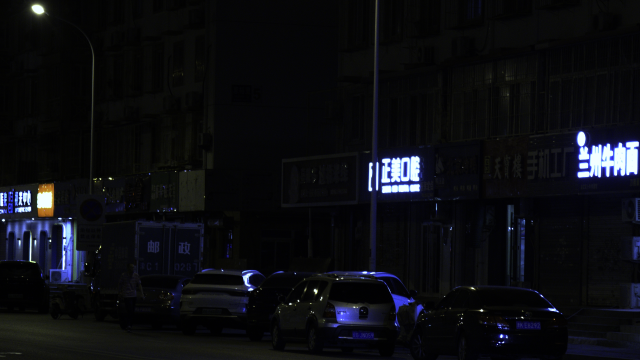} &
\includegraphics[width=0.16\textwidth]{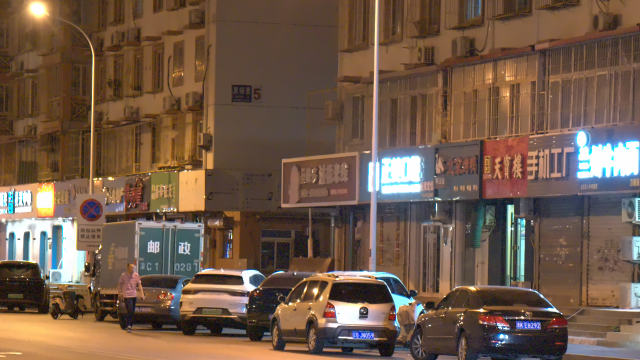} &
\includegraphics[width=0.16\textwidth]{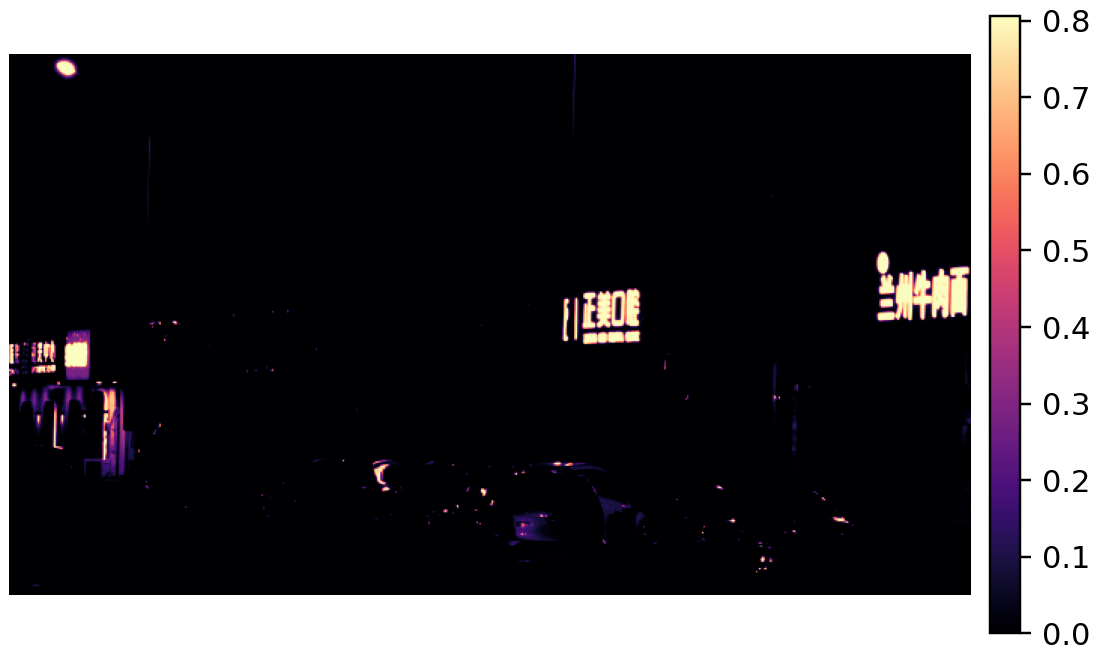} &
\includegraphics[width=0.13\textwidth]{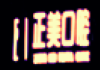} &
\includegraphics[width=0.28\textwidth]{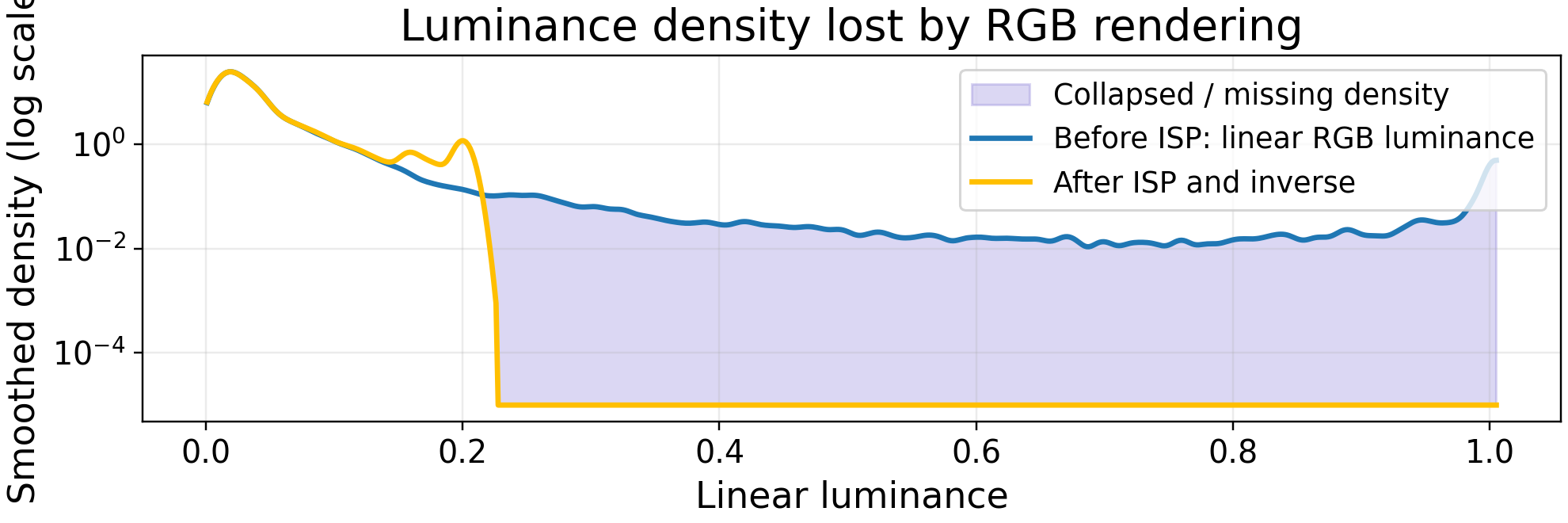} \\
\midrule
\includegraphics[width=0.16\textwidth]{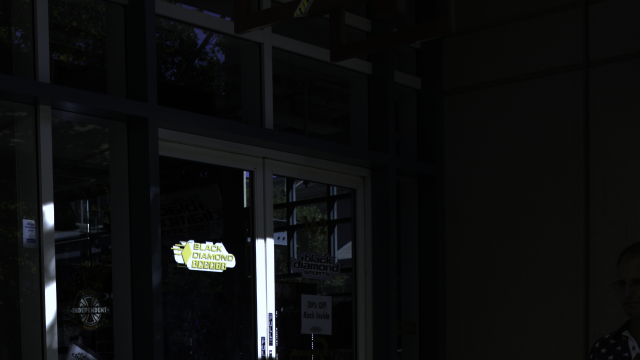} &
\includegraphics[width=0.16\textwidth]{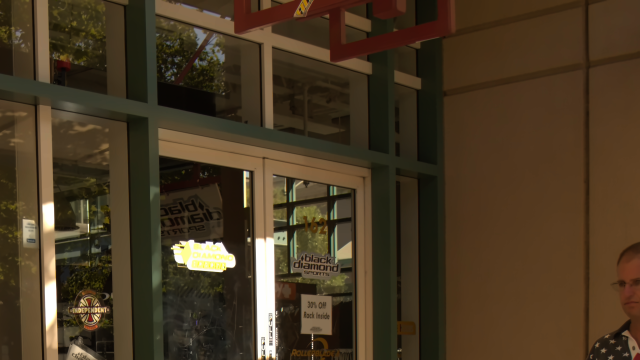} &
\includegraphics[width=0.16\textwidth]{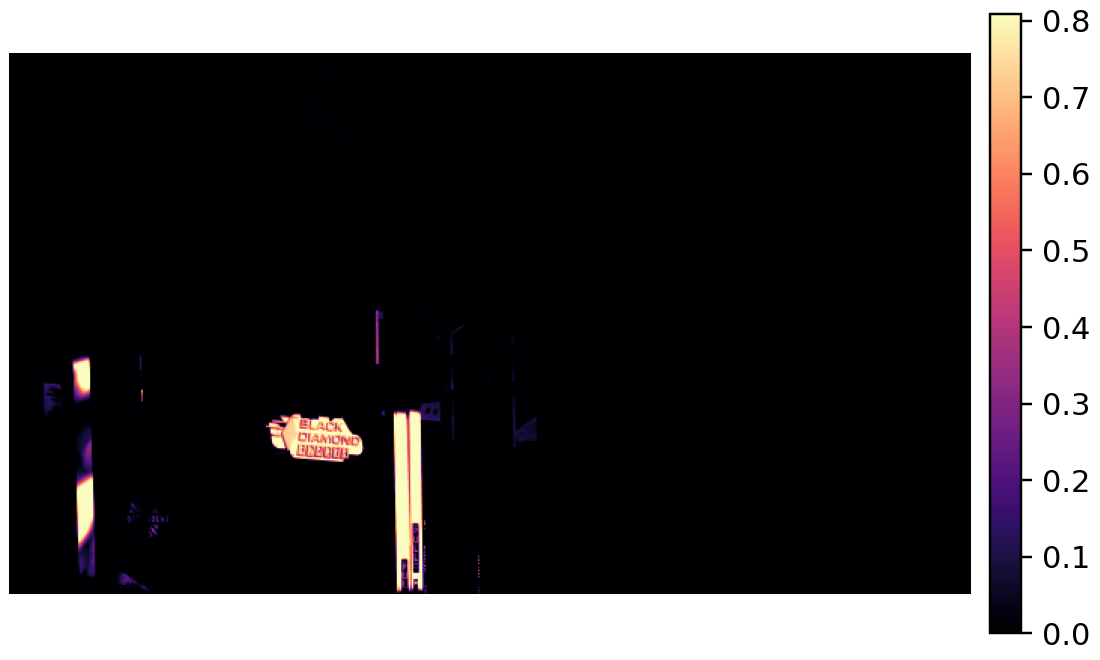} &
\includegraphics[width=0.13\textwidth]{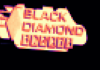} &
\includegraphics[width=0.28\textwidth]{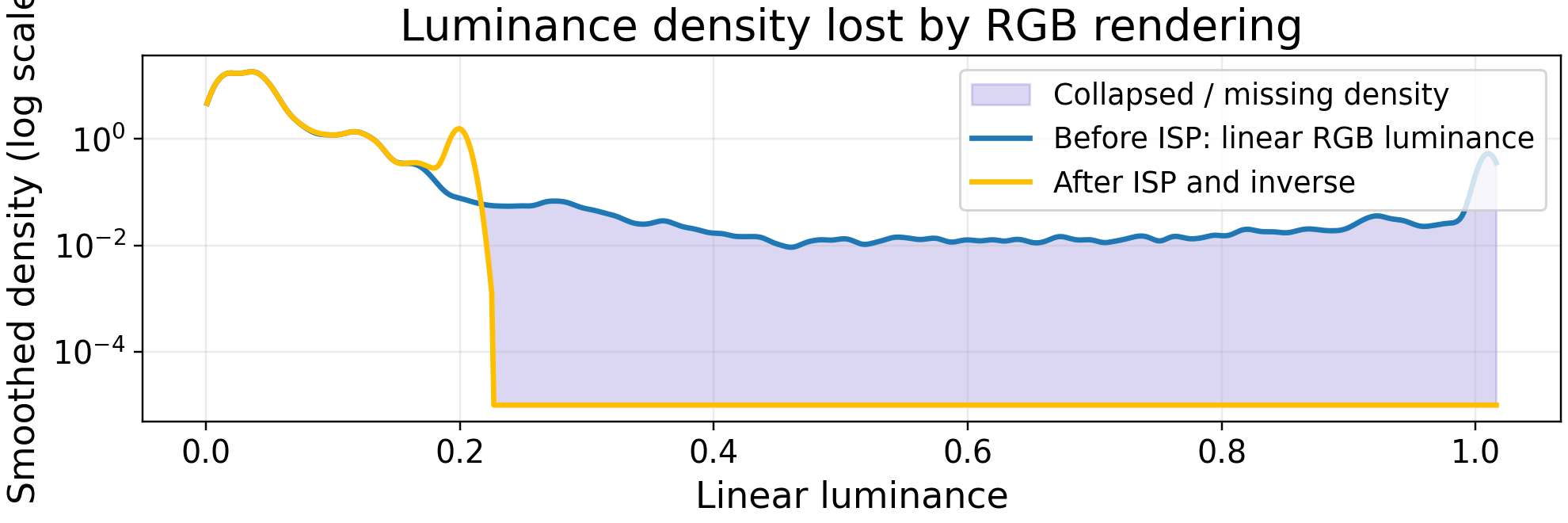} \\
\bottomrule
\end{tabular}%
}
\caption{\textbf{Conventional RGB rendering can discard task-critical visual evidence.} For each example, we compare the Meas.-XYZ observation, its conventionally rendered RGB view, the signal that cannot be recovered after RGB rendering and inverse processing, a local crop of this lost-signal residual, and the corresponding luminance distribution. In the fourth column, the blue regions mark visual signal that is present in the measurement-domain observation but lost after RGB rendering and inverse recovery. The residual and crop show that this missing signal concentrates on the illuminated text regions needed to answer the question, indicating that the RGB image no longer preserves key measurement-domain evidence for grounding.}
\label{fig:isp_inverse_loss}
\end{figure*}

For many everyday scenes, the distinction between measurement and rendering is harmless. In low-light, back-lit, high-dynamic-range (HDR), and exposure-imbalanced scenes, it can be decisive. Weak text strokes may be denoised away, highlights may saturate, shadow structure may be compressed, and noise statistics may be reshaped before a model sees any pixels. In such cases, a VLM failure can originate upstream of reasoning: the model may not be hallucinating over sufficient evidence, but reasoning from an observation interface that has already removed the evidence needed to answer. Figure~\ref{fig:isp_inverse_loss} illustrates this failure mode: clipped RGB rendering collapses part of the measurement-domain signal on illuminated text regions, making inverse recovery insufficient.

This paper studies the observation interface as a first-class design variable for multimodal grounding. Rather than asking whether RAW is a better image format in isolation, we ask whether a VLM can ground answers more reliably when it operates on a measurement-domain input and interprets that representation under camera capture context. We call this setting \emph{measurement-grounded vision-language learning}. Throughout the paper, RAW denotes the source capture or provenance, measurement-domain input denotes the model-facing representation derived from sensor measurements, and Meas.-XYZ denotes the concrete measurement-domain input used in our implementation.

This setting is not solved by replacing an RGB rendering with an unprocessed RAW tensor. Three obstacles arise. First, the model must consume a sensor-proximal measurement-domain representation that preserves linear measurement evidence while remaining compatible with a VLM visual encoder. Second, instruction supervision is easiest to obtain from human-viewable RGB renderings, so supervision must be transferred from appearance space back to measurement-domain examples. Third, measurement-domain evidence is not context-free: ISO, exposure time, aperture, and related capture variables affect how latent visual signals should be interpreted. We therefore formulate the problem around a RAW-to-measurement transformation, a proxy-based supervision-generation process, and camera-conditioned grounding.

We instantiate this formulation as \emph{PRISM-VL}. PRISM-VL uses a measurement-domain representation as the model-facing visual input, constructs instruction supervision through BracketSup, and injects capture metadata through both metadata-augmented questions and residual conditioning in late visual layers. This design mirrors the causal chain of the failure mode: preserve evidence before rendering discards it, transfer supervision from the space where annotation is reliable to the space where learning occurs, and condition interpretation on the physics of capture.

The formulation yields a falsifiable empirical signature. If rendered RGB is a bottleneck, measurement grounding should not merely improve a single cherry-picked slice; it should produce a broad but nonuniform right-shift in grounding quality, with the largest gains in regimes where rendering is least faithful to the underlying evidence. Our benchmark and ablations are organized around this prediction: the main comparisons characterize the RGB-native baseline landscape under the benchmark protocol, while controlled PRISM variants isolate the roles of measurement-domain input, camera conditioning, and BracketSup.

Our contributions are as follows:
\begin{itemize}[leftmargin=1.5em]
    \item We identify the visual \emph{observation interface} as a source of VLM grounding error: rendered RGB can discard measurement evidence before inference begins.
    \item We formulate and instantiate \emph{measurement-grounded vision-language learning} as PRISM-VL, which combines Meas.-XYZ inputs, camera-conditioned grounding, and Exposure-Bracketed Supervision Aggregation (BracketSup).
    \item We construct a quality-controlled 150K instruction-tuning resource and a held-out benchmark covering low-light, HDR, visibility-sensitive, and hallucination-sensitive grounding cases.
    \item Empirically, PRISM-VL-8B reaches 0.6120 BLEU, 0.4571 ROUGE-L, and 82.66\% LLM-Judge accuracy, improving over the RGB Qwen3-VL-8B baseline by +0.1074 BLEU, +0.1071 ROUGE-L, and +4.46\%.
\end{itemize}

\begin{figure*}[h]
\centering
\includegraphics[width=0.99\textwidth]{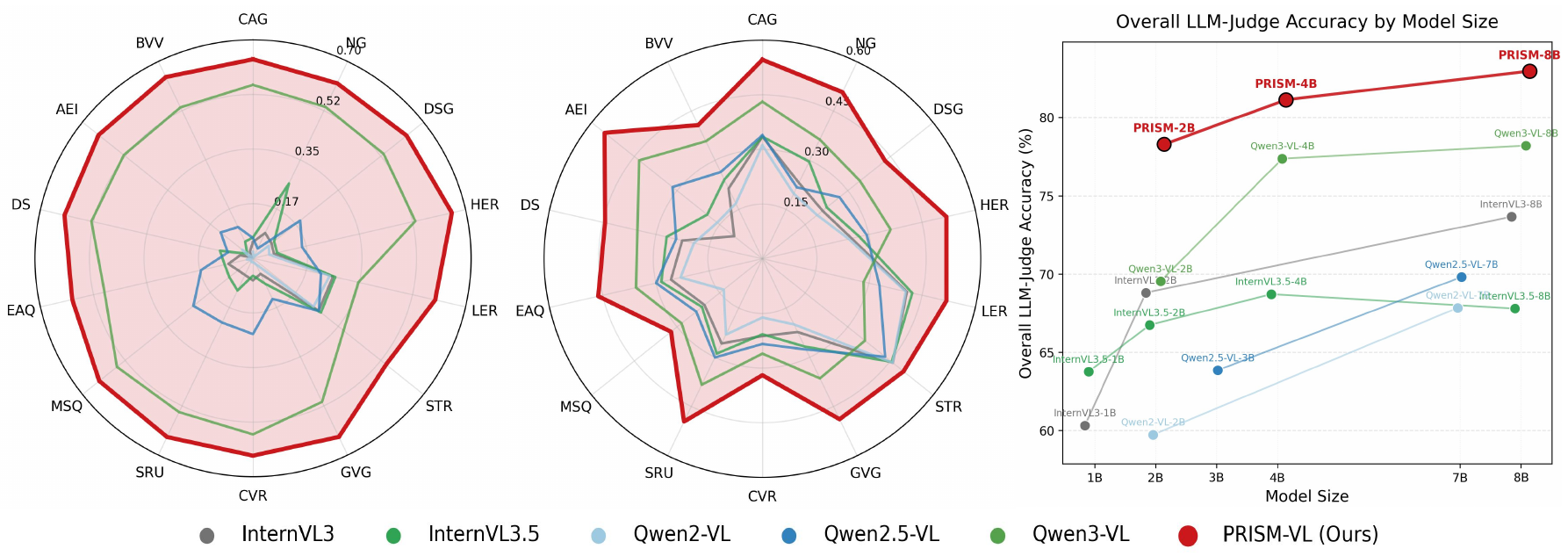}
\caption{
\textbf{A capability fingerprint of measurement grounding.}
The radar reports LLM-Judge accuracy over the benchmark capability dimensions: chromatic attribute grounding (CAG), numerosity grounding (NG), descriptive scene grounding (DSG), HDR evidence recovery (HER), low-illumination evidence recovery (LER), scene text recognition (STR), general visual grounding (GVG), compositional visual reasoning (CVR), spatial relation understanding (SRU), manner and state queries (MSQ), entity and attribute queries (EAQ), discriminative selection (DS), agent and entity identification (AEI), and binary visual verification (BVV). Larger radius indicates higher accuracy. Compared with representative RGB-native VLM families at their strongest evaluated scales, PRISM-VL-8B yields a broader capability profile, highlighting where measurement-domain evidence most changes grounding behavior.
}
\label{fig:capability_radar}
\end{figure*}

\section{Related Work}
\label{sec:related}

\paragraph{RGB-Native Vision-Language Modeling.}
Contemporary VLMs are built largely on post-ISP RGB corpora and RGB-pretrained visual encoders \citep{radford2021clip,alayrac2022flamingo,li2023blip2,liu2023llava,wang2024qwen2vl,Bai2025Qwen25VLTR,Bai2025Qwen3VLTR,Zhu2025InternVL3EA,Wang2025InternVL35AO}. This recipe has enabled rapid progress in open-world multimodal understanding, but it also makes rendered RGB the de facto observation interface for language grounding. Our work does not dispute the utility of RGB-native VLMs or their scaling trajectory. Instead, it isolates a premise that is usually implicit: the rendered image is treated as a sufficient carrier of visual evidence. We study the regimes where this premise is weakest, especially when grounding depends on weak, clipped, or exposure-sensitive signals.

\paragraph{Instruction Data and Benchmark Construction for Multimodal Systems.}
Large-scale instruction data \citep{liu2023llava,dai2023instructblip,li2023m3it,chen2023sharegpt4v} and diagnostic benchmarks \citep{liu2023mmbench,fu2023mme,yue2023mmmu} have become central to VLM progress. This line of work establishes that model behavior depends strongly on how supervision is elicited, filtered, and evaluated. In the measurement-domain setting, the supervision problem becomes more constrained: current annotators are strongest on human-viewable RGB renderings, whereas the target learner should operate on a sensor-derived input rather than the rendered proxy. PRISM-VL therefore treats data construction as part of the method. Multi-exposure proxy annotation is not merely a dataset scaling trick; it is an appearance-to-measurement transfer operator that converts reliable appearance-space annotations into supervision attached to the underlying RAW capture.

Taken together, prior work establishes two premises that motivate our study: RGB-native VLMs define rendered images as the dominant interface for multimodal grounding, and instruction-data research shows that supervision and evaluation protocols strongly shape VLM behavior. PRISM-VL connects these premises through a focused question: what changes when the interface between physical measurement and multimodal reasoning is moved closer to sensor evidence than to rendered appearance, while supervision remains constructed through reliable appearance-space annotations?

\section{Measurement-Grounded Vision-Language Learning}
\label{sec:data}

\subsection{Problem Formulation}
\label{sec:formulation}

A visual interface determines which physical evidence is available before language reasoning begins. RGB-native VLMs implicitly choose post-ISP rendered images as this interface; measurement-grounded VLMs instead make the interface explicit and treat it as part of the learning problem. This separates three design choices that are often conflated in system comparisons: the observation seen by the model, the interface used to obtain supervision, and the capture context needed to interpret measured evidence.

Let a RAW capture be denoted by $x$, its capture metadata by $m$, and its instruction record by $y=(q,a)$. We define
\begin{equation}
    z=\mathcal{T}(x), \qquad
    y=\mathcal{G}(x), \qquad
    \bar{q}=\kappa(q,m),
\end{equation}
where $\mathcal{T}$ maps the capture into a model-facing measurement-domain observation, $\mathcal{G}$ constructs supervision tied to the same capture, and $\kappa$ serializes capture context into the query. The target model is written as
\begin{equation}
    p_{\theta}(a \mid z, \bar{q}, m),
\end{equation}
with $\bar{q}=q$ when question-side conditioning is disabled. In this paper, $z$ is instantiated as \emph{Meas.-XYZ}: a normalized RAW-derived linear XYZ representation. The remaining occurrence of $m$ denotes structured visual-side metadata conditioning when enabled.

\begin{figure*}[t]
\centering
\includegraphics[width=0.9\textwidth]{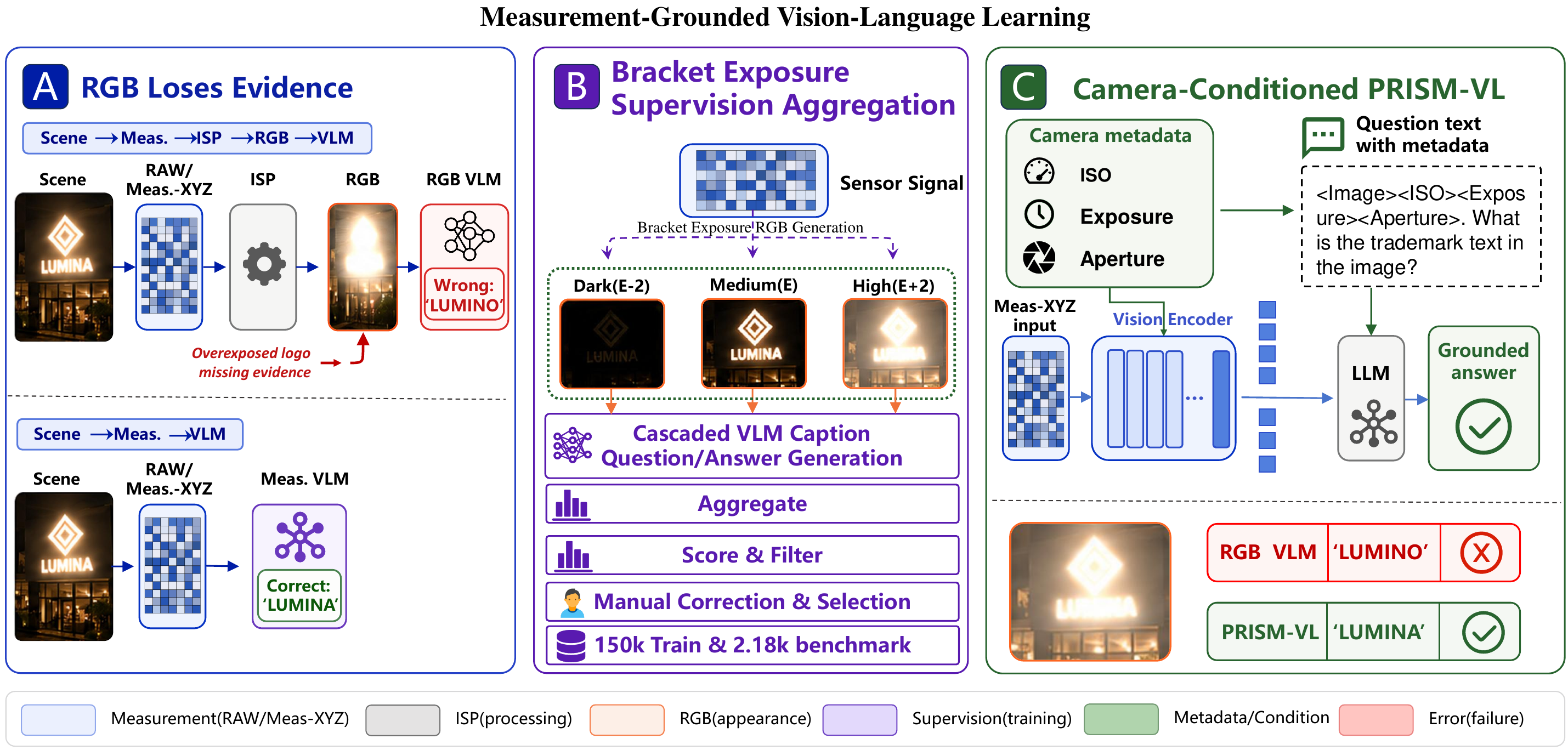}
\caption{
\textbf{Measurement-grounded vision-language learning from sensor evidence.}
RGB-native VLMs reason over ISP-rendered RGB images whose visual information has already been partially lost before inference begins. PRISM-VL instead trains and evaluates on Meas.-XYZ observations, uses the Bracket Exposure Supervision Aggregation module in panel B to construct supervision from exposure-bracketed RGB proxies, and conditions grounding on camera metadata through language-side and visual-side pathways.
}
\label{fig:method_overview}
\end{figure*}

Figure~\ref{fig:method_overview} summarizes the resulting separation between observation, supervision, and conditioning. RGB renderings are used as annotation proxies, while training and inference operate on the measurement-domain observation.

\subsection{Observation and Supervision Interfaces}
\label{sec:meas_xyz_interface}
\label{sec:supervision_transfer}

Measurement grounding distinguishes the observation interface from the annotation interface. The observation operator $\mathcal{T}$ produces the tensor seen by the model. We instantiate it as Meas.-XYZ, which is not the RAW mosaic itself but a dense, three-channel, linear measurement-domain representation expressed in XYZ space. This preserves a closer relation to sensor evidence than post-ISP sRGB while remaining compatible with standard VLM visual tokenization.

The supervision operator $\mathcal{G}$ is needed because current multimodal annotators are more reliable on human-viewable renderings than on measurement-domain inputs. We therefore use exposure-conditioned RGB proxies only at annotation time:
\begin{equation}
    \mathcal{G}(x)=\Gamma(\{\mathcal{A}(\rho_e(x))\}_{e\in\mathcal{E}}),
\end{equation}
where $\rho_e$ renders an exposure-conditioned proxy, $\mathcal{A}$ produces candidate supervision, and $\Gamma$ aggregates the candidates into one instruction record. The resulting sample pairs the record $y=(q,a)$ with the measurement-domain observation $z$, so proxy renderings transfer supervision rather than replace measurement-domain learning.

Given a training set $\mathcal{D}$ built from these paired records, PRISM-style models optimize the standard autoregressive objective
\begin{equation}
    \mathcal{L}(\theta)=
    \mathbb{E}_{(z,q,a,m)\sim\mathcal{D}}
    \left[-\log p_{\theta}\!\left(a\mid z,\bar{q},m\right)\right].
\end{equation}
This factorization lets later experiments isolate whether gains arise from the observation interface $\mathcal{T}$, camera conditioning, or the supervision-transfer operator $\mathcal{G}$.

\subsection{Measurement-Domain Data Construction}
\label{sec:measurement_data_construction}

We instantiate the supervision operator $\mathcal{G}$ from RAW captures in RAISEDNG \citep{dangnguyen2015raise}, AODDNG \citep{li2024aodraw}, and PASCALRAW \citep{omidzohoor2015pascalraw}. Because current multimodal annotators are better calibrated to human-viewable RGB than to linear measurement tensors, we annotate RGB proxies and transfer the supervision to the corresponding Meas.-XYZ inputs. The pipeline performs captioning, question generation and post-processing, answer generation, and QA scoring; BracketSup further aggregates exposure-conditioned proxies from brighter or darker renderings tied to the same RAW example.

We then filter and balance the records before instruction tuning. Starting from roughly 700K auto-annotated candidates, we retain 518,433 post-scoring records, remove placeholder-answer failures, and balance source prefixes, question types, repeated templates, and score floors. The final split contains 150,000 Meas.-XYZ instruction examples with capture metadata.

\subsection{Held-Out Benchmark Design}
\label{sec:heldout_benchmark_design}

The held-out benchmark isolates observation-interface failures under low-light, HDR, visibility-sensitive, and hallucination-sensitive conditions, while retaining RGB-sufficient cases for breadth. Table~\ref{tab:benchmark_dimensions_main} summarizes the multi-dimensional capability taxonomy; rows denote evaluated grounding behaviors rather than internal data buckets.

\begin{table}[h]
\centering
\caption{Capability dimensions in the held-out multi-dimensional benchmark.}
\label{tab:benchmark_dimensions_main}
\scriptsize
\setlength{\tabcolsep}{3pt}
\renewcommand{\arraystretch}{1.05}
\begin{tabular}{lll}
\toprule
Abbrev. & Capability Dimension & Evaluation Focus \\
\midrule
CAG & Chromatic Attribute Grounding & Colors and chromatic attributes. \\
NG & Numerosity Grounding & Object or instance counts. \\
DSG & Descriptive Scene Grounding & Scene-level grounded descriptions. \\
HER & HDR Evidence Recovery & Evidence in high dynamic range scenes. \\
LER & Low-Illumination Evidence Recovery & Weak evidence under poor illumination. \\
STR & Scene Text Recognition & Standard scene text. \\
GVG & General Visual Grounding & General visual queries. \\
CVR & Compositional Visual Reasoning & Entities, attributes, and relations. \\
SRU & Spatial Relation Understanding & Layouts and relative positions. \\
MSQ & Manner and State Queries & Actions, conditions, and visual states. \\
EAQ & Entity and Attribute Queries & Entities and their attributes. \\
DS & Discriminative Selection & Candidate selection. \\
AEI & Agent and Entity Identification & People, agents, or salient entities. \\
BVV & Binary Visual Verification & Yes/no visual propositions. \\
\bottomrule
\end{tabular}
\end{table}

For controlled RGB-versus-Meas.-XYZ comparisons, benchmark captures are held out before paired views are materialized. The split has zero sample-index overlap and disjoint RAW paths from training; where metadata permit, we also separate scene, device, and capture session. The final benchmark contains 2,183 matched examples after preprocessing.

\subsection{Camera-Conditioned Grounding}

Measurement-domain evidence is not self-interpreting: the same measured signal can correspond to different visual conditions under different exposure, aperture, or ISO settings. We therefore treat capture metadata as part of the grounding context rather than as incidental dataset bookkeeping. The conditioning map $\kappa(q,m)$ exposes this context on the language side, while structured use of $m$ allows the visual pathway to interpret weak, clipped, or noisy evidence under the conditions in which it was captured. Section~\ref{sec:method} instantiates these two conditioning paths in PRISM-VL, and Section~\ref{sec:experiments} evaluates their contribution with matched ablations.

\section{PRISM-VL: Instantiating Measurement Grounding}
\label{sec:method}

PRISM-VL implements the formulation above with three components: Meas.-XYZ observations, camera-conditioned grounding, and BracketSup instruction tuning. The method keeps the backbone VLM interface close to Qwen3-VL-Instruct, but changes which visual evidence reaches the model, how capture context conditions that evidence, and how supervision is attached to measurement-domain inputs.

\subsection{Architecture Overview}

A RAW capture is transformed by $\mathcal{T}$ into the Meas.-XYZ observation $z$, which replaces rendered RGB as the visual input. The visual encoder and projector produce image tokens for the LLM as in the base VLM, while capture metadata enters through a language-side question context and a visual-side residual conditioning path. Supervision comes from $\mathcal{G}$, which attaches a consolidated instruction record to the same underlying capture. PRISM-VL is therefore an intervention on the evidence interface and conditioning pathway rather than a new general-purpose VLM backbone.

\subsection{Meas.-XYZ Observation Interface}

Meas.-XYZ instantiates $\mathcal{T}$ as a RAW-derived linear XYZ observation. Its role is to preserve measurement structure that can be attenuated by rendering, while maintaining the dense three-channel form expected by existing visual processors. We use it as a controlled replacement for sRGB: the model receives Meas.-XYZ at both training and inference time, and RGB proxies are used only during supervision construction.

\subsection{Camera Conditioning}

PRISM-VL uses camera metadata through two complementary paths. The question-side path implements $\kappa(q,m)$ by appending capture context such as ISO, exposure time, and aperture to the query. This keeps metadata visible to the language model without introducing a separate metadata-token vocabulary.

The visual-side path conditions late visual representations on the same capture context. Let $g(m)$ be a learned projection of normalized metadata into the visual hidden dimension. For selected late visual layers, PRISM-VL applies residual conditioning
\begin{equation}
    h^{(\ell+1)} = \mathrm{Block}_{\ell}(h^{(\ell)}) + g(m), \qquad \ell \in \mathcal{L}_{\mathrm{meta}}.
\end{equation}
Late injection gives the model a structured way to reinterpret semantic visual evidence under capture conditions without forcing early visual filters to become metadata-specific.

\subsection{BracketSup Instruction Tuning}

BracketSup, short for Exposure-Bracketed Supervision Aggregation, instantiates $\mathcal{G}$. For each capture, multiple exposure-conditioned RGB proxies reveal complementary appearance evidence; candidate instruction records from these proxies are aggregated into one supervision record attached to the Meas.-XYZ sample. The proxies are annotation instruments, not training inputs. This makes BracketSup an appearance-to-measurement transfer mechanism: it lets supervision benefit from human-viewable renderings while the learned model remains measurement-grounded.

\begin{table*}[h]
\centering
\caption{Multi-dimensional benchmark comparison between PRISM-VL-2B and RGB Qwen3-VL baselines. Each row corresponds to one capability dimension in the held-out benchmark, and each cell reports BLEU / ROUGE-L. The first column names the evaluated grounding capability rather than the raw dataset bucket.}
\label{tab:prism_category_comparison}
\setlength{\tabcolsep}{2.5pt}
\resizebox{\textwidth}{!}{%
\begin{tabular}{l c c c c}
\toprule
Capability Dimension & \small{Qwen3-VL-2B} & \small{Qwen3-VL-4B} & \small{Qwen3-VL-8B} & \small{PRISM-VL-2B} \\
 & \small{BLEU / ROUGE-L} & \small{BLEU / ROUGE-L} & \small{BLEU / ROUGE-L} & \small{BLEU / ROUGE-L} \\
\midrule
Chromatic Attribute Grounding (CAG) & 0.2445 / 0.3751 & 0.3344 / 0.3922 & 0.5557 / 0.4312 & \textbf{0.6043} / \textbf{0.4980} \\
Numerosity Grounding (NG) & 0.4546 / 0.3387 & 0.4759 / 0.3447 & 0.5379 / 0.3633 & \textbf{0.5980} / \textbf{0.4538} \\
Descriptive Scene Grounding (DSG) & 0.3926 / 0.2928 & 0.4735 / 0.3408 & 0.5365 / 0.3429 & \textbf{0.5986} / \textbf{0.3901} \\
HDR Evidence Recovery (HER) & 0.3712 / 0.3179 & 0.4663 / 0.3548 & 0.5343 / 0.3614 & \textbf{0.6066} / \textbf{0.4533} \\
Low-Illumination Evidence Recovery (LER) & 0.3051 / 0.3368 & 0.3443 / 0.3064 & 0.3470 / 0.2851 & \textbf{0.5174} / \textbf{0.4249} \\
Scene Text Recognition (STR) & 0.3491 / 0.4040 & 0.3847 / 0.4094 & 0.3719 / 0.3604 & \textbf{0.5084} / \textbf{0.4669} \\
General Visual Grounding (GVG) & 0.3970 / 0.3315 & 0.4736 / 0.3557 & 0.5109 / 0.3644 & \textbf{0.6117} / \textbf{0.4505} \\
Compositional Visual Reasoning (CVR) & 0.3345 / 0.2223 & 0.5430 / 0.2739 & 0.5646 / 0.2603 & \textbf{0.6052} / \textbf{0.2944} \\
Spatial Relation Understanding (SRU) & 0.2713 / 0.3217 & 0.3791 / 0.3421 & 0.5472 / 0.3836 & \textbf{0.6093} / \textbf{0.4740} \\
Manner and State Queries (MSQ) & 0.3004 / 0.2333 & 0.5114 / 0.2800 & 0.5585 / 0.2841 & \textbf{0.5946} / \textbf{0.2876} \\
Entity and Attribute Queries (EAQ) & 0.3271 / 0.3255 & 0.4464 / 0.3536 & 0.4889 / 0.3560 & \textbf{0.5741} / \textbf{0.4397} \\
Discriminative Selection (DS) & 0.3201 / 0.2971 & 0.4869 / 0.3412 & 0.5319 / 0.3522 & \textbf{0.5820} / \textbf{0.4047} \\
Agent and Entity Identification (AEI) & 0.4063 / 0.3451 & 0.4208 / 0.4161 & 0.5304 / 0.4332 & \textbf{0.6210} / \textbf{0.5307} \\
Binary Visual Verification (BVV) & 0.3149 / 0.2862 & 0.5341 / 0.3457 & 0.5367 / 0.3580 & \textbf{0.6186} / \textbf{0.3732} \\
\bottomrule
\end{tabular}%
}
\end{table*}

\section{Experiments and Analysis}
\label{sec:experiments}

\begin{table*}[h]
\centering
\caption{Overall performance of representative RGB baselines and PRISM-VL variants on the benchmark. Judge values are percentages.}
\label{tab:all_methods_overall}
\scriptsize
\setlength{\tabcolsep}{3.2pt}
\resizebox{\textwidth}{!}{%
\begin{tabular}{l *{11}{c} *{3}{c}}
\toprule
Metric &
\multicolumn{2}{c}{InternVL3} &
\multicolumn{2}{c}{InternVL3.5} &
\multicolumn{2}{c}{Qwen2-VL} &
\multicolumn{2}{c}{Qwen2.5-VL} &
\multicolumn{3}{c}{Qwen3-VL} &
\multicolumn{3}{c}{\textbf{PRISM-VL}} \\
\cmidrule(lr){2-3}\cmidrule(lr){4-5}\cmidrule(lr){6-7}
\cmidrule(lr){8-9}\cmidrule(lr){10-12}\cmidrule(l){13-15}
& 2B & 8B
& 4B & 8B
& 2B & 7B
& 3B & 7B
& 2B & 4B & 8B
& \textbf{2B} & \textbf{4B} & \textbf{8B} \\
\midrule
BLEU &
0.0638 & 0.0938 & 0.1951 & 0.1265 &
0.0595 & 0.0627 & 0.0775 & 0.1671 & 0.3407 & 0.4442 & 0.5046 &
\textbf{0.5865} & \textbf{0.6021} & \textbf{0.6120} \\
ROUGE-L &
0.2458 & 0.2621 & 0.3109 & 0.2891 &
0.2328 & 0.2412 & 0.2537 & 0.2908 & 0.3171 & 0.3453 & 0.3500 &
\textbf{0.4244} & \textbf{0.4465} & \textbf{0.4571} \\
Judge (\%) &
68.80 & 73.66 & 68.71 & 67.80 &
59.73 & 67.84 & 63.86 & 69.81 & 69.54 & 77.37 & 78.20 &
\textbf{77.99} & \textbf{80.83} & \textbf{82.66} \\
\bottomrule
\end{tabular}%
}
\end{table*}

\subsection{Experimental Setup}

We evaluate three questions: whether measurement-domain input improves grounding over RGB, whether camera conditioning adds benefit, and whether BracketSup improves exposure-sensitive supervision transfer. Together, these experiments separate the primary observation-interface effect from component-level contributions under a shared held-out benchmark. We report BLEU, ROUGE-L, and LLM-Judge accuracy, using overall comparisons for the main result and matched ablations for attribution.

\subsection{Main Results: Is RGB the Right Observation Interface?}

Table~\ref{tab:prism_category_comparison} tests whether changing the observation interface from rendered RGB to measurement-domain input improves grounding across capability dimensions. PRISM-VL-2B outperforms RGB Qwen3-VL baselines on BLEU and ROUGE-L across all reported dimensions, indicating that the gain is not concentrated in a single subset. Table~\ref{tab:all_methods_overall} extends the comparison to the broader RGB-native baseline landscape and to PRISM-VL models at 2B, 4B, and 8B scales.

The overall results show a substantial ranking shift under the benchmark protocol. PRISM-VL-2B surpasses RGB Qwen3-VL-8B on BLEU and ROUGE-L while nearly matching its LLM-Judge accuracy with a smaller backbone, and scaling the same recipe to PRISM-VL-8B reaches 0.6120 BLEU, 0.4571 ROUGE-L, and 82.66\% LLM-Judge accuracy. These comparisons establish the empirical effect relative to RGB-native VLMs; Tables~\ref{tab:bracketsup_category_ablation}--\ref{tab:component_ablation} then separate dataset-specific fine-tuning, camera conditioning, and BracketSup from the broader measurement-grounding pipeline.

\begin{table}[h]
\centering
\caption{Effect of BracketSup on exposure-sensitive capability dimensions. We compare matched PRISM-VL-2B variants with and without Exposure-Bracketed Supervision Aggregation, reporting LLM-Judge accuracy on Low-Illumination Evidence Recovery (LER) and HDR Evidence Recovery (HER). BracketSup improves both dimensions, with the larger gain in LER where single-render supervision is least reliable.}
\label{tab:bracketsup_category_ablation}
\scriptsize
\setlength{\tabcolsep}{4pt}
\begin{tabular}{l c c}
\toprule
Variant & LER & HER \\
\midrule
W/o BracketSup & 55.49\% & 76.37\% \\
+ BracketSup & \textbf{67.08\%} & \textbf{79.70\%} \\
\bottomrule
\end{tabular}
\end{table}

\begin{table}[h]
\centering
\caption{Compact controls for dataset-specific fine-tuning and metadata-value intervention. Left: RGB-ft fine-tunes Qwen3-VL backbones on the RGB version of the same instruction data used by PRISM-VL. Right: the same camera-conditioned PRISM-VL variant is evaluated with real, zeroed, and shuffled metadata over a 280-example intervention set. Values report LLM-Judge accuracy.}
\label{tab:matched_rgb_finetuning}
\label{tab:metadata_value_intervention}
\scriptsize
\setlength{\tabcolsep}{4pt}
\begin{tabular}{l c c c | l c}
\toprule
\multicolumn{4}{c|}{Matched RGB fine-tuning control} & \multicolumn{2}{c}{Metadata value intervention} \\
\cmidrule(r){1-4}\cmidrule(l){5-6}
Model Size & RGB zero-shot & RGB-ft & PRISM-VL & Setting & LLM-Judge \\
\midrule
2B & 69.54\% & 74.16\% & \textbf{77.99\%} & Real meta & \textbf{78.13\%} \\
4B & 77.37\% & 77.60\% & \textbf{80.83\%} & Zero meta & 77.06\% \\
8B & 78.20\% & 78.84\% & \textbf{82.66\%} & Shuffled meta & 75.27\% \\
\bottomrule
\end{tabular}%
\end{table}

\begin{table}[h]
\centering
\caption{Component ablation of PRISM-VL-2B. The three component columns indicate question-side metadata conditioning, visual residual metadata conditioning, and Exposure-Bracketed Supervision Aggregation (BracketSup). Adding each component improves or preserves BLEU, ROUGE-L, and LLM-Judge accuracy, with the full model achieving the strongest overall performance.}
\label{tab:component_ablation}
\scriptsize
\setlength{\tabcolsep}{3pt}
\begin{tabular}{c c c c c c}
\toprule
Question metadata & Visual metadata & BracketSup & BLEU & ROUGE-L & LLM-Judge \\
\midrule
-- & -- & -- & 0.5580 & 0.3914 & 74.48\% \\
\checkmark & -- & -- & 0.5748 & 0.3982 & 74.78\% \\
\checkmark & \checkmark & -- & 0.5784 & 0.4076 & 74.95\% \\
\checkmark & \checkmark & \checkmark & \textbf{0.5865} & \textbf{0.4244} & \textbf{77.99\%} \\
\bottomrule
\end{tabular}%
\end{table}

PRISM-VL-2B improves the overall LLM-Judge accuracy from 70.16\% to 77.99\%, corresponding to a 7.83 percentage-point absolute gain and an 11.16\% relative gain over the RGB baseline.

\subsection{Matched RGB Fine-Tuning Control}

Because PRISM-VL is trained on our constructed instruction data, a natural question is whether its improvement comes from dataset-specific fine-tuning rather than from the measurement-domain observation itself. The left block of Table~\ref{tab:matched_rgb_finetuning} addresses this by fine-tuning matched Qwen3-VL backbones on the RGB version of the same instruction data and evaluating them under the same benchmark. RGB fine-tuning improves the 2B, 4B, and 8B RGB baselines to 74.16\%, 77.60\%, and 78.84\% LLM-Judge accuracy, respectively, confirming that the constructed data is useful. However, PRISM-VL remains higher at every model scale. Since the data source, backbone family, and evaluation protocol are matched, the remaining gap is attributable to the observation signal: Meas.-XYZ preserves task-relevant measurement evidence that the rendered RGB input has already lost or compressed.

\subsection{Ablation of Camera-Conditioned Grounding}

Table~\ref{tab:component_ablation} shows that camera metadata gives modest gains on top of Meas.-XYZ: question-side metadata raises LLM-Judge accuracy from 74.18\% to 74.48\%, visual residual conditioning reaches 74.65\%, and the full BracketSup model reaches 77.99\% with the best BLEU and ROUGE-L scores. The 280-example intervention in Table~\ref{tab:metadata_value_intervention} further shows that real metadata outperforms zeroed and shuffled values by 1.07 and 2.86 points, indicating that metadata is useful context but not the dominant source of improvement.

\subsection{Ablation of Exposure-Bracketed Supervision Transfer}

Table~\ref{tab:bracketsup_category_ablation} isolates BracketSup using matched Qwen3-VL-2B variants with the same question-side and residual metadata conditioning. Adding Exposure-Bracketed Supervision Aggregation raises LER LLM-Judge accuracy from 55.49\% to 67.08\% and HER from 76.37\% to 79.70\%, supporting BracketSup as an appearance-to-measurement transfer mechanism for weak or exposure-sensitive evidence.

\begin{table*}[!t]
\centering
\scriptsize
\setlength{\tabcolsep}{3pt}
\renewcommand{\arraystretch}{1.15}
\caption{\textbf{Qualitative comparison on Low-Illumination Evidence Recovery (LER) examples with weak text evidence.} We compare RGB and Meas.-XYZ observations with evidence heatmaps, answer-region crops, and model responses. RGB Qwen3-VL grounds on incorrect text, whereas PRISM-VL recovers the reference answers from measurement-domain evidence.}
\label{tab:exocr_qualitative_examples}
\resizebox{0.9\textwidth}{!}{%
\begin{tabular}{p{0.10\textwidth} p{0.14\textwidth} p{0.14\textwidth} p{0.14\textwidth} p{0.14\textwidth} p{0.14\textwidth} p{0.14\textwidth}}
\toprule
\multicolumn{7}{c}{\textbf{Visual Input Example}} \\
\midrule
Task query & \multicolumn{3}{p{0.42\textwidth}}{What is the name of the illuminated shop next to the Beijing Roast Duck?} & \multicolumn{3}{p{0.42\textwidth}}{What is the word on the first line of the yellow sign?} \\
\midrule
Reference answer & \multicolumn{3}{p{0.42\textwidth}}{\textcolor{green!35!black}{\zhengmeid (Zhengmei Dental Clinic)}} & \multicolumn{3}{p{0.42\textwidth}}{\textcolor{green!35!black}{BLACK}} \\
\midrule
 & Image & Evidence heatmap & Zoomed crop & Image & Evidence heatmap & Zoomed crop \\
RGB observation &
\includegraphics[width=\linewidth]{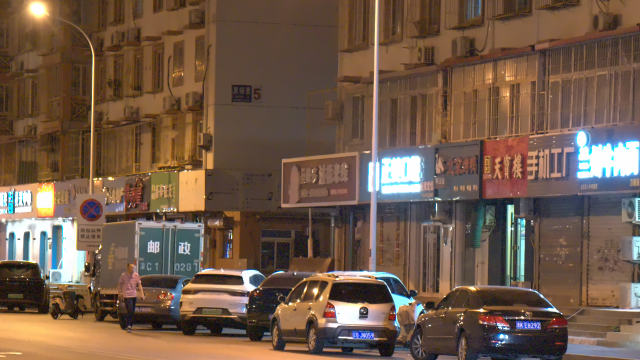} &
\includegraphics[width=\linewidth]{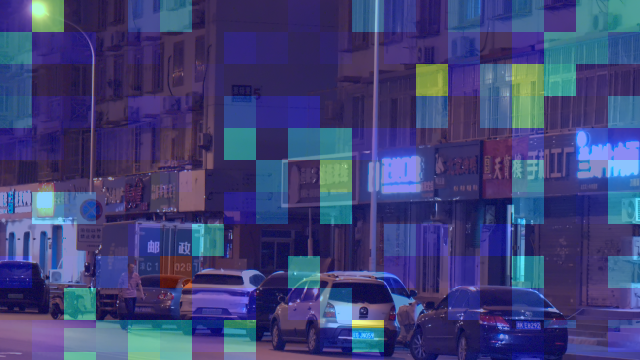} &
\includegraphics[width=\linewidth]{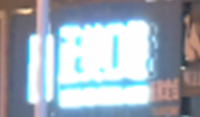} &
\includegraphics[width=\linewidth]{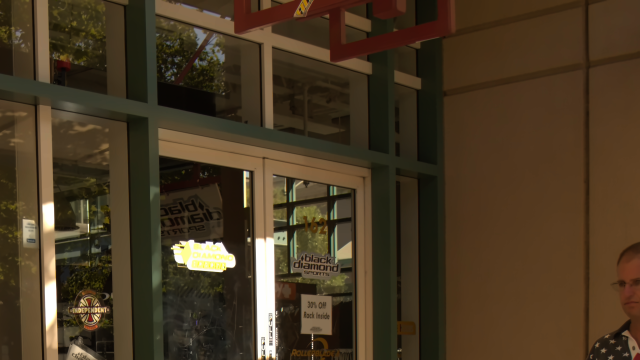} &
\includegraphics[width=\linewidth]{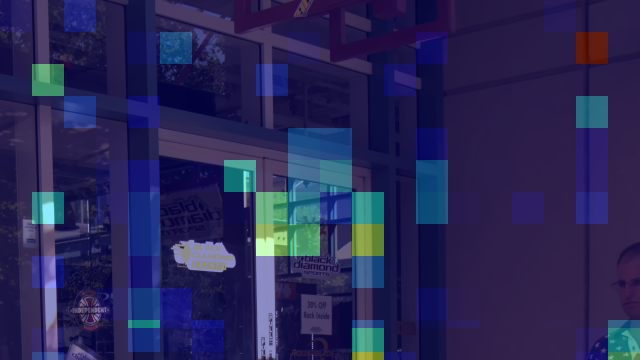} &
\includegraphics[width=\linewidth]{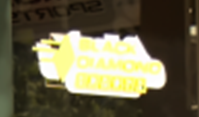} \\
\midrule
RGB Qwen3-VL answer & \multicolumn{3}{p{0.42\textwidth}}{The illuminated shop next to the "Beijing Roast Duck" is the one with the sign that says \textcolor{red}{"Hua Tian Hua" (\huatianhua)}. \badmark} & \multicolumn{3}{p{0.42\textwidth}}{The first line of the yellow sign is \textcolor{red}{\emph{diamond}}. \badmark} \\
\midrule
 & Image & Evidence heatmap & Zoomed crop & Image & Evidence heatmap & Zoomed crop \\
Meas.-XYZ observation &
\includegraphics[width=\linewidth]{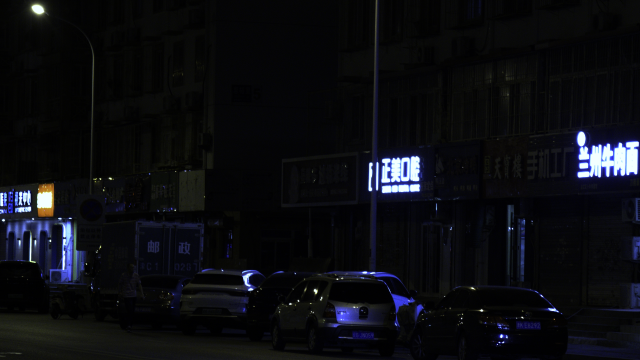} &
\includegraphics[width=\linewidth]{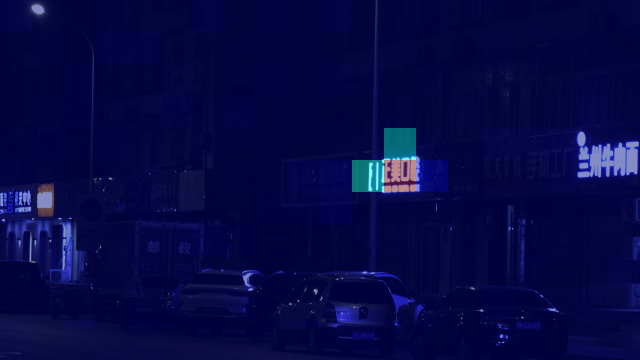} &
\includegraphics[width=\linewidth]{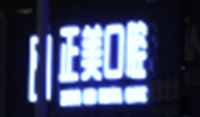} &
\includegraphics[width=\linewidth]{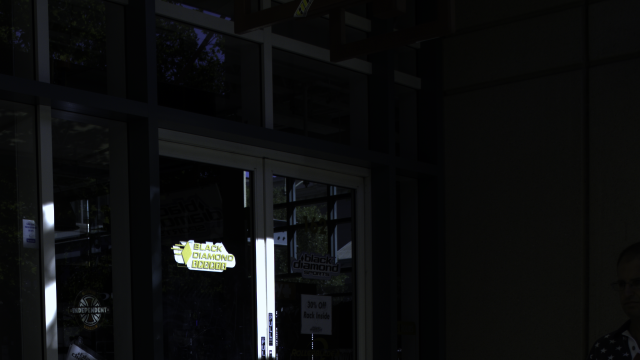} &
\includegraphics[width=\linewidth]{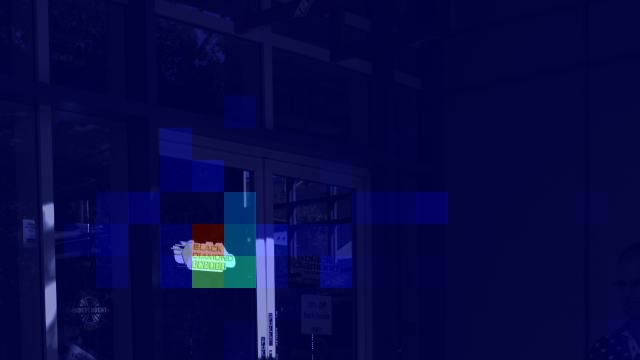} &
\includegraphics[width=\linewidth]{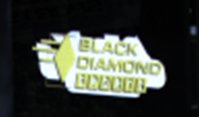} \\
\midrule
PRISM-VL answer & \multicolumn{3}{p{0.42\textwidth}}{The illuminated shop next to \beijingduck (Beijing Roast Duck) is named \textcolor{green!35!black}{\textbf{\zhengmeid} (Zhengmei Dental Clinic)}. \goodmark} & \multicolumn{3}{p{0.42\textwidth}}{The first line of the yellow sign is \textcolor{green!35!black}{\textbf{BLACK}}. \goodmark} \\
\bottomrule
\end{tabular}%
}
\end{table*}

\subsection{Qualitative Analysis}

Table~\ref{tab:exocr_qualitative_examples} shows two Low-Illumination Evidence Recovery (LER) examples where RGB Qwen3-VL grounds on incorrect weak text, while PRISM-VL recovers the reference answer from measurement-domain evidence. The comparison illustrates that the gain is not only numerical: RGB and Meas.-XYZ expose different evidence to the model.

Figure~\ref{fig:isp_inverse_loss} further isolates the rendering failure. We render the measurement-domain image through an ISP-style RGB path with exposure gain, clipping, sRGB transfer, and 8-bit quantization, then invert this rendering and measure the unrecoverable residual. The lost signal concentrates on illuminated text regions; under the analyzed setting, 5.79\% and 3.20\% of pixels are clipped in the two examples, and the recovered 99th-percentile luminance is capped at 0.20 versus 0.96 and 0.98 in the original linear signal. Thus, the relevant evidence is not merely reparameterized by RGB rendering, but partially discarded before VLM reasoning begins.

\FloatBarrier
\section{Conclusion}
\label{sec:conclusion}

We introduced measurement-grounded vision-language learning and instantiated it as PRISM-VL, combining Meas.-XYZ observations, camera-conditioned grounding, and BracketSup supervision transfer. Across physically challenging scenes, PRISM-VL shows that preserving measurement-domain evidence enables more reliable grounding than relying on rendered RGB alone. These results suggest that future VLMs should treat the visual observation interface, not only model scale and data volume, as a central design choice.



\end{document}